\documentclass{article}
\usepackage[preprint]{neurips_2026}
\makeatletter
\renewcommand{\@noticestring}{}
\makeatother

\usepackage[utf8]{inputenc}
\usepackage[T1]{fontenc}
\usepackage{hyperref}
\usepackage{url}
\usepackage{booktabs}
\usepackage{amsmath,amsfonts,amssymb}
\usepackage{microtype}
\usepackage{xcolor}
\usepackage{graphicx}
\usepackage{subcaption}
\usepackage{multirow}
\usepackage{array}
\usepackage{algorithm}
\usepackage{algpseudocode}
\usepackage{tikz}
\usetikzlibrary{positioning,arrows.meta}
\usepackage{enumitem}

\title{Beyond Demographic Balance: Multi-Metric and Intersectional Evaluation of Fairness in MIMIC-IV Mortality Prediction}

\author{
Abdullah Al Noman\\
Virginia Tech\\
Blacksburg, Virginia, USA\\
\texttt{noman@vt.edu}
\And
Fahmid Al Rifat\\
University of Texas at Arlington\\
Arlington, Texas, USA\\
\texttt{fahmidal.rifat@uta.edu}
\AND
Tahrima Hashem\\
University of Melbourne\\
Melbourne, Victoria, Australia\\
\texttt{tahrimacsedu14@gmail.com}
\And
Syed Muhammad Ibne Zulfiker\\
University of Toronto\\
Toronto, Ontario, Canada\\
\texttt{syed.ibnezulfiker@utoronto.ca}
\AND
Rishov Paul\\
University of Virginia\\
Charlottesville, Virginia, USA\\
\texttt{vst2hb@virginia.edu}
\And
Tanzima Hashem\\
Bangladesh University of Engineering and Technology\\
Dhaka, Bangladesh\\
\texttt{tanzimahashem@cse.buet.ac.bd}
}
\begin{document}
\maketitle

\begin{abstract}
Fairness conclusions in clinical prediction can depend strongly on both the metrics reported and the demographic resolution at which performance is evaluated. We revisit these evaluation choices for ICU mortality prediction on MIMIC-IV, comparing predictive-utility and subgroup-error metrics across several fairness interventions. As a complementary case study, we introduce a lightweight adaptation strategy that jointly balances ethnicity--gender--insurance representation without conditioning on mortality outcomes, allowing demographic representation balancing to be examined separately from outcome-conditioned or direct error-rate interventions. We evaluate its behavior at both marginal and corresponding three-way intersectional subgroup levels, while accounting for the statistical support of finer-grained estimates. The results show that interventions can receive substantially different assessments across accuracy/AUROC, sensitivity, and false-positive rate, and that marginal demographic summaries can conceal heterogeneous error profiles within their constituent intersections, including among larger subgroups. These findings highlight the importance of evaluating fairness interventions at both complementary metric and subgroup resolutions, while accounting for the intervention target and the reliability of subgroup estimates.
\end{abstract}

\section{Introduction}

Machine-learning models trained on electronic health records (EHRs) can exhibit substantial performance differences across demographic groups because EHR data reflect variation in disease burden, access to care, measurement processes, coding practices, and representation in the training population \citep{obermeyer2019racial,celi2022sources,chen2023algorithmic}. Such disparities are consequential in clinical risk prediction because false negatives and false positives have different implications for patients and health-system resources. Aggregate accuracy or discrimination therefore provides only partial evidence about whether a model behaves comparably across patient groups.

This creates an evaluation problem as well as a mitigation problem. Different metrics capture different properties: AUROC summarizes ranking across thresholds, whereas sensitivity (i.e., true-positive rate) and false-positive rate characterize errors at a particular operating point. Likewise, a marginal analysis by ethnicity, gender, or insurance averages over the remaining attributes, while \textbf{intersectional analysis} evaluates \emph{combinations of these attributes} and can reveal subgroup-specific errors that marginal summaries conceal. Fairness conclusions can therefore change with both \emph{metric choice} and \emph{subgroup resolution}. When outcome prevalence differs across demographic groups, no single statistical fairness criterion is universally appropriate, and calibration and error-rate parity cannot generally be satisfied simultaneously \citep{chouldechova2017fair,pleiss2017fairness}. Recent clinical work similarly argues that evaluation should reflect the errors and subgroup structure relevant to the decision context rather than be inferred from demographic parity alone \citep{lett2025intersectional}.

We use this distinction to study a simple intervention on MIMIC-IV ICU mortality prediction \citep{johnson2023mimiciv}. The predictor is a time-series LSTM built on the preprocessing pipeline of \citet{gupta2022pipeline}. Motivated by lightweight last-layer adaptation \citep{mao2023lastlayer}, we form a fine-tuning cohort by jointly stratifying ethnicity, gender, and insurance and downsampling every observed intersection to the smallest subgroup. We freeze the learned representation and optimize a replacement classifier head on this cohort. Importantly, the mortality label is \emph{not} part of the balancing criterion. The intervention therefore targets demographic representation without prescribing equal mortality prevalence across intersections or directly optimizing subgroup error parity.

Prior work considers demographic--outcome balancing and outcome-conditioned augmentation \citep{mao2023lastlayer,theodorou2025fairplay}. Here, we instead exclude the ``mortality'' outcome from the sampling criterion to isolate the effect of changing demographic representation during adaptation. This distinction is relevant in clinical data because mortality prevalence can differ across demographic subgroups, and conditioning the balancing procedure on outcome would change both demographic representation and output-label composition. 
The resulting design provides a focused case study for examining what fairness conclusions are supported when adaptation acts on demographic representation alone.

However, multiple protected attributes make fairness evaluation more challenging. Marginal analysis can conceal intersection-specific errors, while finer stratification reduces subgroup and positive-outcome counts, increasing the tension between demographic resolution and statistical support. Prior work has highlighted the importance of multi-attribute and intersectional fairness in general machine learning and healthcare \citep{deng2023orthogonal,zhu2023m3fair,chen2024multiple,lett2025intersectional}. Complementing these studies, we jointly balance ethnicity--gender--insurance representation during head adaptation without conditioning on mortality outcomes, and evaluate the resulting predictions at both marginal and three-way intersectional resolutions. We emphasize better-supported subgroups and account for uncertainty to support reliable interpretation of the results. To summarize, we address the following two research questions:

\begin{enumerate}[nosep, leftmargin=*]
    \item \textbf{RQ1: Metric dependence.} Do predictive-utility and subgroup-error metrics support consistent conclusions about interventions when demographic representation is balanced without conditioning on mortality outcomes?
    \item \textbf{RQ2: Evaluation resolution.} To what extent do marginal demographic evaluations conceal error behavior at the corresponding ethnicity--gender--insurance intersections?
\end{enumerate}

Our contributions are threefold. First, we provide empirical evidence that fairness conclusions in MIMIC-IV mortality prediction depend on the choice of predictive-utility and subgroup-error metrics. Second, we quantify how marginal evaluation can obscure heterogeneity across ethnicity--gender--insurance intersections. Third, we study a lightweight head-adaptation strategy that jointly balances demographic intersections without conditioning on mortality outcomes, enabling demographic representation balancing to be evaluated separately from outcome-conditioned or error-rate-based interventions.

\begin{table}[t]
\centering
\small
\setlength{\tabcolsep}{3.5pt}
\begin{tabular}{p{0.17\textwidth}p{0.19\textwidth}p{0.28\textwidth}p{0.28\textwidth}}
\toprule
Work & Data/task & Intervention target & Main distinction \\
\midrule

Mao et al. \citep{mao2023lastlayer}
& Vision benchmarks
& Sensitive group $\times$ label; optional fairness regularization
& Last-layer adaptation outside clinical EHRs \\
\addlinespace[2pt]

Lett et al. \citep{lett2025intersectional}
& MIMIC-IV ED admission
& Ethnoracial identity $\times$ gender performance objectives
& Direct intersectional performance optimization; different endpoint \\
\addlinespace[2pt]

FairPlay \citep{theodorou2025fairplay}
& MIMIC-IV/eICU mortality
& Demographic--outcome-conditioned augmentation
& Synthetic records with explicit outcome conditioning \\
\addlinespace[2pt]

Yoon and Kwak \citep{yoon2026singleaxis}
& MIMIC-IV mortality
& Race or gender targeted separately
& Single-axis interventions with cross-axis/intersectional evaluation \\

This study
& MIMIC-IV ICU mortality
& Ethnicity $\times$ gender $\times$ insurance representation
& Joint three-way balancing of observed records without outcome conditioning \\
\bottomrule
\end{tabular}
\caption{Closest methodological contrasts in intervention target and demographic resolution.}
\label{tab:related}
\end{table}

\section{Related Work}
\label{sec:related}

\paragraph{Fairness in clinical prediction.}
Bias in clinical machine learning can arise from data collection, labels, representation, model development, and downstream use \citep{obermeyer2019racial,celi2022sources,chen2023algorithmic}. On MIMIC-IV, \citet{meng2022mimicif} reported demographic performance disparities in in-hospital mortality prediction. More broadly, clinical fairness studies vary substantially in cohort construction, endpoints, protected attributes, and evaluation criteria \citep{liu2025scoping}, making the evaluation protocol an important part of how fairness results should be interpreted.

\paragraph{Outcome prevalence and fairness criteria.}
Demographic balancing, demographic--outcome balancing, and direct error-rate optimization target different aspects of the prediction problem. When outcome prevalence differs across groups, calibration and equalized-error criteria cannot generally be achieved simultaneously \citep{chouldechova2017fair,pleiss2017fairness}. In MIMIC-IV emergency-admission prediction, \citet{lett2025intersectional} therefore emphasize performance-based fairness criteria over requiring equal positive prediction rates. FairPlay takes a different approach by conditioning synthetic augmentation on both demographic factors and clinical outcomes \citep{theodorou2025fairplay}. In contrast, our intervention modifies demographic representation during adaptation without conditioning the balancing rule on mortality outcomes.

\paragraph{Multiple sensitive attributes and intersectional fairness.}
Prior work has examined multiple sensitive attributes through orthogonal representation learning in medical imaging \citep{deng2023orthogonal} and multi-attribute reweighting in healthcare data \citep{zhu2023m3fair}. On MIMIC-IV, \citet{lett2025intersectional} compare marginal and intersectional debiasing, while \citet{yoon2026singleaxis} show that race- or gender-targeted interventions in mortality prediction can produce cross-axis effects and metric-dependent conclusions. Related empirical work likewise shows that fairness interventions can behave differently across protected attributes and evaluation metrics \citep{chen2024multiple}. Our setting complements these studies by balancing ethnicity, gender, and insurance jointly and evaluating the resulting behavior at both marginal and corresponding three-way intersectional resolutions.

\paragraph{Lightweight classifier adaptation.}
\citet{mao2023lastlayer} improve fairness by adapting the final classifier of a pretrained network using sensitive-group--label balanced data, with variants that include explicit fairness penalties. Our configuration instead balances three demographic attributes jointly without including the mortality label and adapts a small replacement MLP head while keeping the learned representation fixed. Thus, the intervention targets demographic representation during adaptation rather than demographic--outcome balance or direct error-rate constraints. Table~\ref{tab:related} summarizes the closest methodological contrasts.

\section{Methods}
\label{sec:methods}

\subsection{Prediction task and base model}

MIMIC-IV is a de-identified critical-care database derived from Beth Israel Deaconess Medical Center \citep{johnson2023mimiciv}. We use the ICU mortality prediction pipeline derived from \citet{gupta2022pipeline}. After preprocessing, the experiments contain 17,883 training stays, including 2,152 mortality-positive stays (12.0\%), and 4,400 held-out test stays, including 494 mortality-positive stays (11.2\%).

The base predictor is a time-series LSTM. Dynamic ICU modalities are embedded and projected into a shared latent representation together with static diagnosis and demographic embeddings. A two-layer LSTM with hidden size 256 encodes the temporal sequence, followed by a $256\!\rightarrow\!128$ fully connected layer and a $128\!\rightarrow\!1$ sigmoid classifier. The reported configuration uses Adam with learning rate $10^{-3}$, batch size 200, and a maximum of 20 epochs. The demographic variables used in the fairness analysis are ethnicity (Asian, Black, Hispanic/Latino, Other, White), gender (female, male), and insurance (Medicaid, Medicare, Other), yielding up to $5\times2\times3=30$ intersections.

\subsection{Intersectionally balanced head fine-tuning}
\label{sec:ibhft}

Let $D_{\mathrm{tr}}=\{(x_i,y_i,a_i)\}_{i=1}^{N}$ denote the training set, where $y_i\in\{0,1\}$ is the mortality label and $a_i=(e_i,g_i,s_i)$ contains ethnicity, gender, and insurance. Let $\mathcal{C}$ be the set of observed demographic intersections and $D_c=\{(x_i,y_i,a_i):a_i=c\}$ the records belonging to intersection $c$. We define
\begin{equation}
    n_{\min}=\min_{c\in\mathcal{C}} |D_c|,
\end{equation}
and construct the fine-tuning cohort
\begin{equation}
    D_{\mathrm{bal}}=\bigcup_{c\in\mathcal{C}} \operatorname{Sample}(D_c,n_{\min}),
\end{equation}
where sampling is without replacement. The label $y$ does not enter the grouping criterion. The intervention therefore enforces equal sample counts across observed demographic intersections,
\begin{equation}
    |D^{\mathrm{bal}}_c| = n_{\min}\quad \forall c\in\mathcal{C},
\end{equation}
but does not impose a target value for $P(Y=1\mid(E,G,I)=c)$ or directly constrain $\mathrm{TPR}_c$ or $\mathrm{FPR}_c$.

The exclusion of $y$ from the balancing rule defines the scope of the intervention. Conditioning additionally on the outcome would introduce a second design choice: balancing across combinations of demographic intersections and mortality labels. Such demographic--outcome balancing has been used in prior work \citep{mao2023lastlayer}, as has outcome-conditioned synthetic augmentation in clinical data \citep{theodorou2025fairplay}. We instead isolate representation balancing and evaluate its consequences. This is particularly relevant when subgroup outcome prevalence differs, because equal representation does not imply equal prevalence or equal error rates \citep{pleiss2017fairness,chouldechova2017fair,lett2025intersectional}. It also avoids requiring a synthetic model of patient records. Under pure downsampling, conditioning on $y$ would further partition up to 30 demographic intersections into as many as 60 demographic--outcome strata; for a relatively infrequent outcome (12.0\% mortality in the training cohort), this can \textit{make the retained sample size more sensitive to rare positive strata.}

The base model is first trained on $D_{\mathrm{tr}}$. For adaptation, all parameters through the 128-dimensional representation are frozen. The original $128\!\rightarrow\!1$ output layer is replaced by a trainable $128\!\rightarrow\!512\!\rightarrow\!256\!\rightarrow\!1$ MLP with ReLU activations and dropout $p=0.3$. Only this classifier head is optimized on $D_{\mathrm{bal}}$ using the binary classification objective. No synthetic records, demographic counterfactuals, or explicit fairness penalty are introduced. Figure~\ref{fig:framework} summarizes the overall study design, linking the intersectionally balanced head adaptation to the subsequent multi-metric and marginal-versus-intersectional evaluation.

\begin{figure}[t]
\centering
\resizebox{0.97\textwidth}{!}{%
\begin{tikzpicture}[
  node distance=6mm and 9mm,
  box/.style={
    draw,
    rounded corners,
    align=center,
    inner sep=5pt,
    minimum height=11mm
  },
  detail/.style={
    draw,
    rounded corners,
    align=center,
    inner sep=4pt,
    minimum height=10mm
  },
  arr/.style={-{Latex[length=2.3mm]}, thick}
]

\node[box] (model)
{Base mortality model\\trained on original cohort};

\node[box, right=of model] (ibhft)
{IBHFT\\intersectionally balanced head fine-tuning};

\node[box, right=of ibhft] (metrics)
{Metric evaluation\\Accuracy/AUROC\\TPR/FPR};

\node[box, right=of metrics] (resolution)
{Demographic resolution\\marginal vs.\ $(E,G,I)$};

\draw[arr] (model) -- (ibhft);
\draw[arr] (ibhft) -- (metrics);
\draw[arr] (metrics) -- (resolution);

\node[detail, below left=8mm and -5mm of ibhft] (balance)
{Intersectional balancing\\
form observed $(E,G,I)$ groups\\
sample $n_{\min}$ records per group};

\node[detail, below right=8mm and -49mm of ibhft] (head)
{Head adaptation\\
freeze 128-D representation\\
train $128\!\rightarrow\!512\!\rightarrow\!256\!\rightarrow\!1$ head};

\draw[arr] (ibhft) -- (balance);
\draw[arr] (ibhft) -- (head);

\end{tikzpicture}%
}
\caption{Study design. IBHFT performs equal-count adaptation over observed ethnicity--gender--insurance intersections and updates only the replacement classifier head. The resulting model is evaluated along two dimensions: predictive-utility versus subgroup-error metrics, and marginal versus intersectional demographic resolution.}
\label{fig:framework}
\end{figure}

\subsection{Comparators and evaluation protocol}

The experimental comparison includes: (i) the unmitigated baseline, (ii) suppression of sensitive attributes, (iii) sample reweighting, (iv) temperature scaling, and (v) intersectionally balanced head fine-tuning (IBHFT). Temperature scaling is included as a calibration control rather than a fairness intervention because it rescales logits without changing their rank ordering. The evaluation protocol varies along two dimensions. First, we compare predictive utility with threshold-dependent subgroup errors. Accuracy and AUROC summarize overall predictive performance, while at threshold 0.5,
\begin{equation}
\mathrm{TPR}=\frac{TP}{TP+FN},\qquad
\mathrm{FPR}=\frac{FP}{FP+TN},\qquad
\mathrm{TNR}=1-\mathrm{FPR}.
\end{equation}
Equalized odds motivates examining TPR and FPR jointly across groups \citep{hardt2016equality}; we therefore do not collapse the evaluation to a single fairness score. RQ1 asks whether these metric families support the same qualitative assessment of the interventions.

Second, we vary demographic resolution. Marginal evaluation considers one attribute at a time, whereas intersectional evaluation uses the full ethnicity--gender--insurance combination. For RQ2, we stratify patient-level predictions by these three attributes and report subgroup size, mortality-positive count ($N_{+}$), sensitivity (TPR), specificity, FPR, accuracy, and AUROC. The main text focuses on better-supported intersections with $N\geq100$ and $N_{+}\geq5$; Appendix~\ref{app:intersectional} reports all intersections with $N_{+}\geq5$, together with 95\% Wilson intervals for sensitivity.

\section{Results}
\label{sec:results}

\subsection{RQ1: metric choice changes the assessment of the interventions}
\begin{figure}[t]
\centering
\begin{subfigure}{0.32\textwidth}\includegraphics[width=\linewidth]{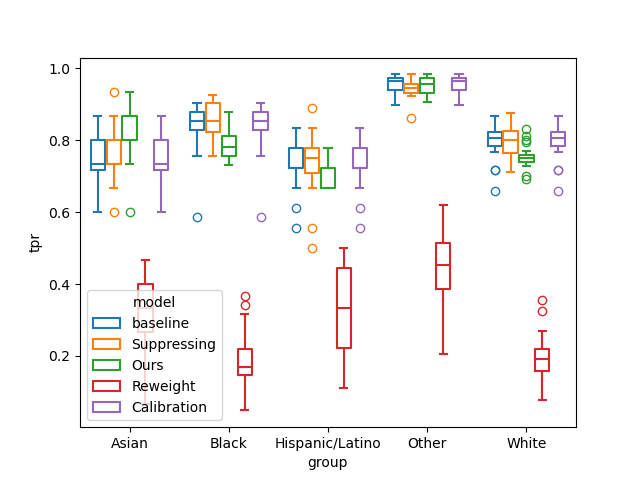}\caption{TPR: ethnicity}\end{subfigure}\hfill
\begin{subfigure}{0.32\textwidth}\includegraphics[width=\linewidth]{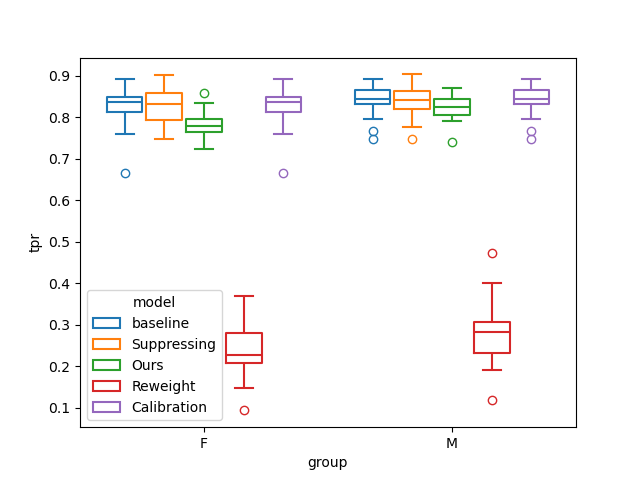}\caption{TPR: gender}\end{subfigure}\hfill
\begin{subfigure}{0.32\textwidth}\includegraphics[width=\linewidth]{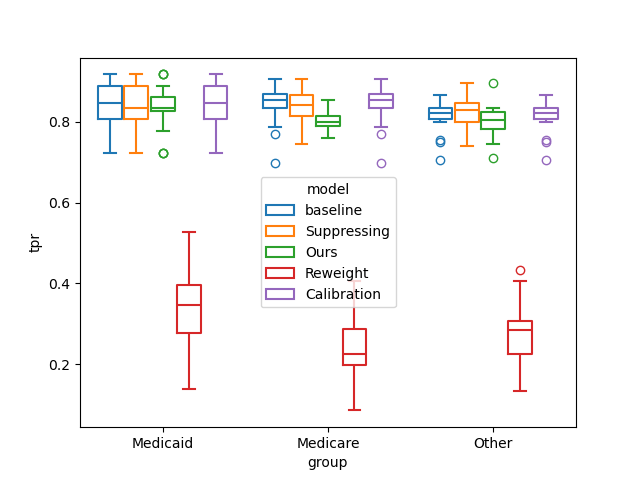}\caption{TPR: insurance}\end{subfigure}

\vspace{-1mm}
\begin{subfigure}{0.32\textwidth}\includegraphics[width=\linewidth]{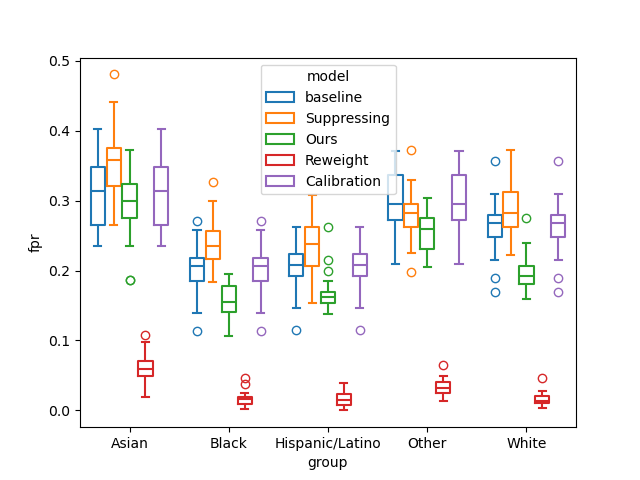}\caption{FPR: ethnicity}\end{subfigure}\hfill
\begin{subfigure}{0.32\textwidth}\includegraphics[width=\linewidth]{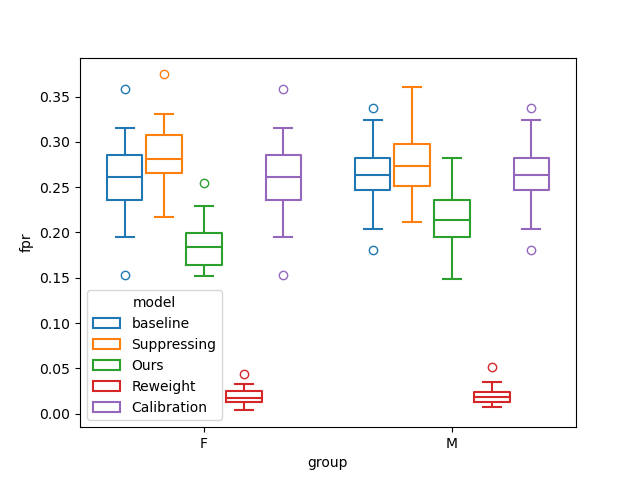}\caption{FPR: gender}\end{subfigure}\hfill
\begin{subfigure}{0.32\textwidth}\includegraphics[width=\linewidth]{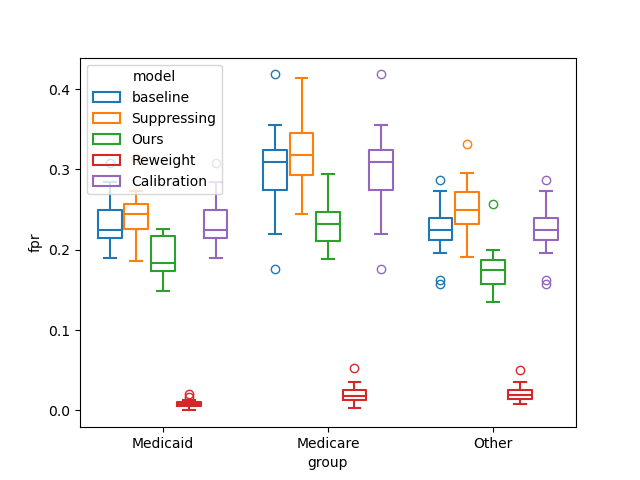}\caption{FPR: insurance}\end{subfigure}
\caption{Group-wise true-positive rate (top) and false-positive rate (bottom) across repeated evaluations. The series labeled ``Ours'' corresponds to IBHFT. Reweighting attains very low FPR but substantially lower TPR, whereas IBHFT preserves a sensitivity profile closer to the baseline.}
\label{fig:tprfpr}
\end{figure}

The predictive-utility and subgroup-error metrics do not support a single, uniform assessment of the evaluated interventions. Figure~\ref{fig:tprfpr} provides evidence for reweighting: although it drives FPR close to zero across ethnicity, gender, and insurance, its TPR drops sharply. Across ethnicity groups, for example, the median TPR is only approximately $0.15$--$0.45$, compared with roughly $0.70$--$0.96$ for the baseline and other interventions. For mortality prediction, the low false-alarm rate therefore reflects a conservative operating profile that misses many true positive cases. Judged by FPR alone, this configuration appears favorable; judged by sensitivity, the conclusion reverses.

IBHFT exhibits a different error profile. Its TPR remains closer to the unmitigated baseline and other comparators, while its FPR is generally reduced relative to the baseline across the marginal demographic groups. For example, across insurance groups, the median FPR decreases from approximately $0.23$ to $0.18$ for Medicaid, $0.31$ to $0.23$ for Medicare, and $0.23$ to $0.18$ for the Other group, while the corresponding TPRs remain around $0.8$ or higher. These results suggest that demographic representation balancing changes the operating characteristics of the classifier, but does not determine a priori which fairness--utility trade-off will result. The effect must therefore be assessed jointly through both false-negative and false-positive behavior.


\begin{figure}[t]
\centering
\begin{subfigure}{0.32\textwidth}\includegraphics[width=\linewidth]{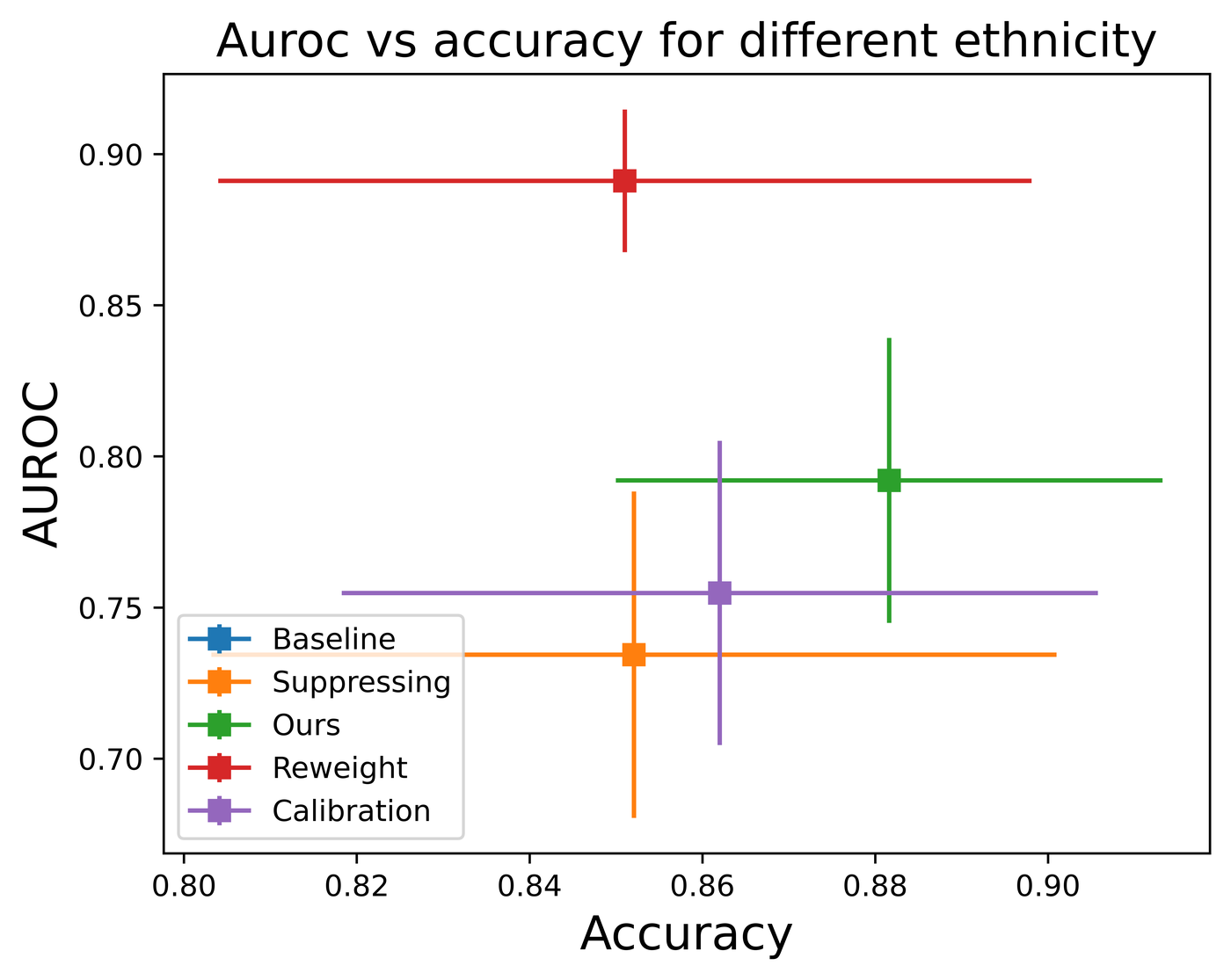}\caption{AUROC vs. accuracy: ethnicity}\end{subfigure}\hfill
\begin{subfigure}{0.32\textwidth}\includegraphics[width=\linewidth]{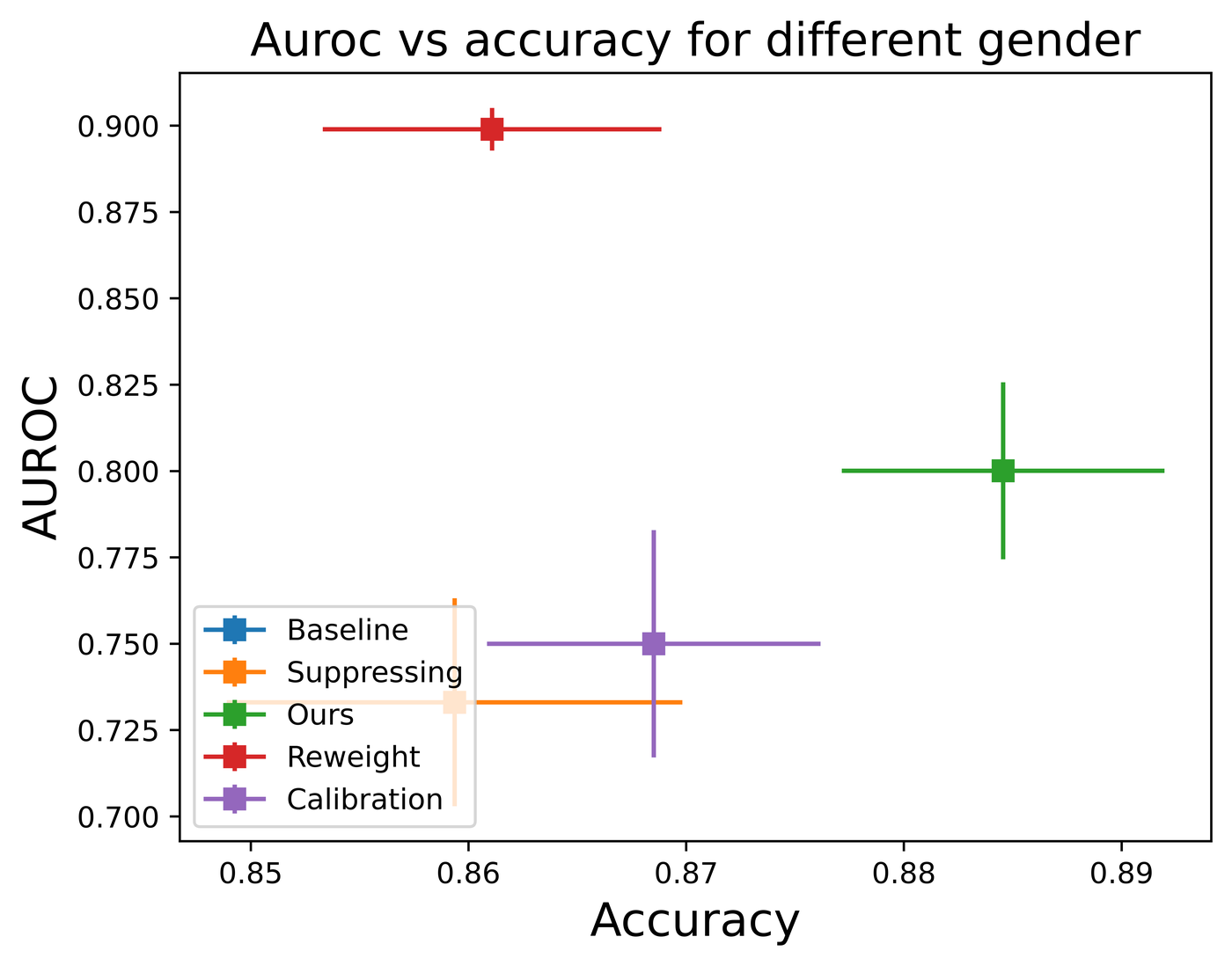}\caption{AUROC vs. accuracy: gender}\end{subfigure}\hfill
\begin{subfigure}{0.32\textwidth}\includegraphics[width=\linewidth]{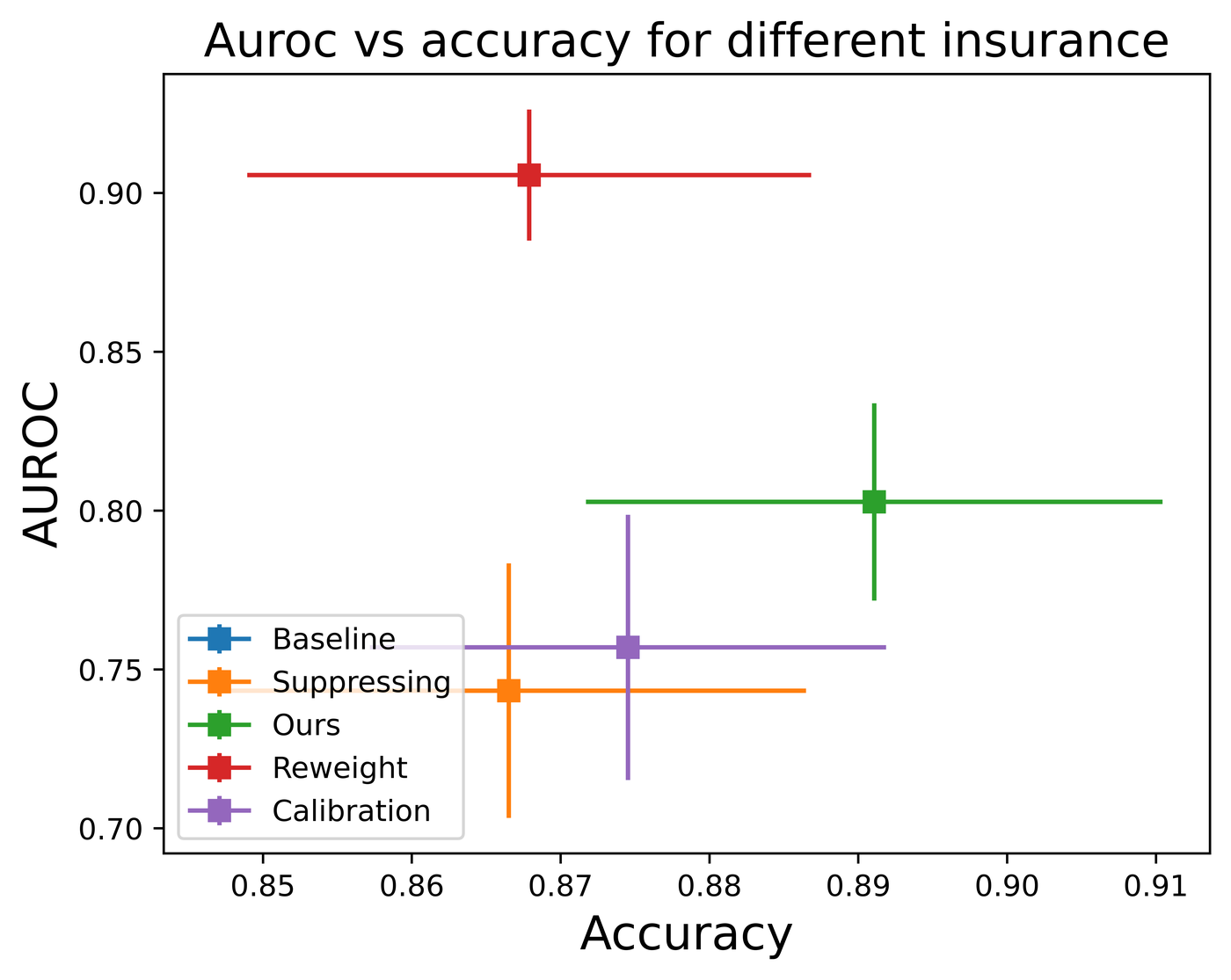}\caption{AUROC vs. accuracy: insurance}\end{subfigure}

\vspace{-1mm}
\begin{subfigure}{0.32\textwidth}\includegraphics[width=\linewidth]{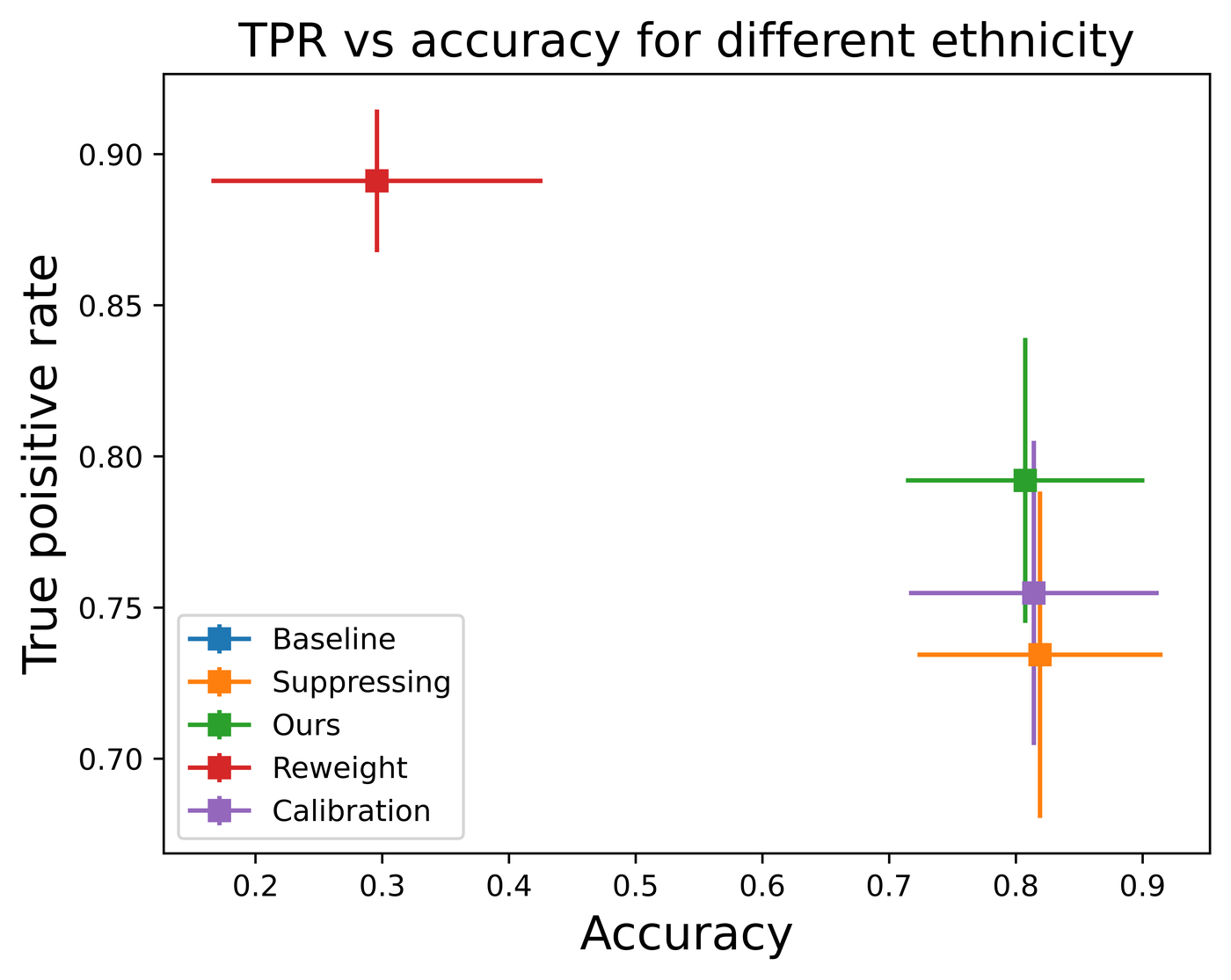}\caption{TPR vs. accuracy: ethnicity}\end{subfigure}\hfill
\begin{subfigure}{0.32\textwidth}\includegraphics[width=\linewidth]{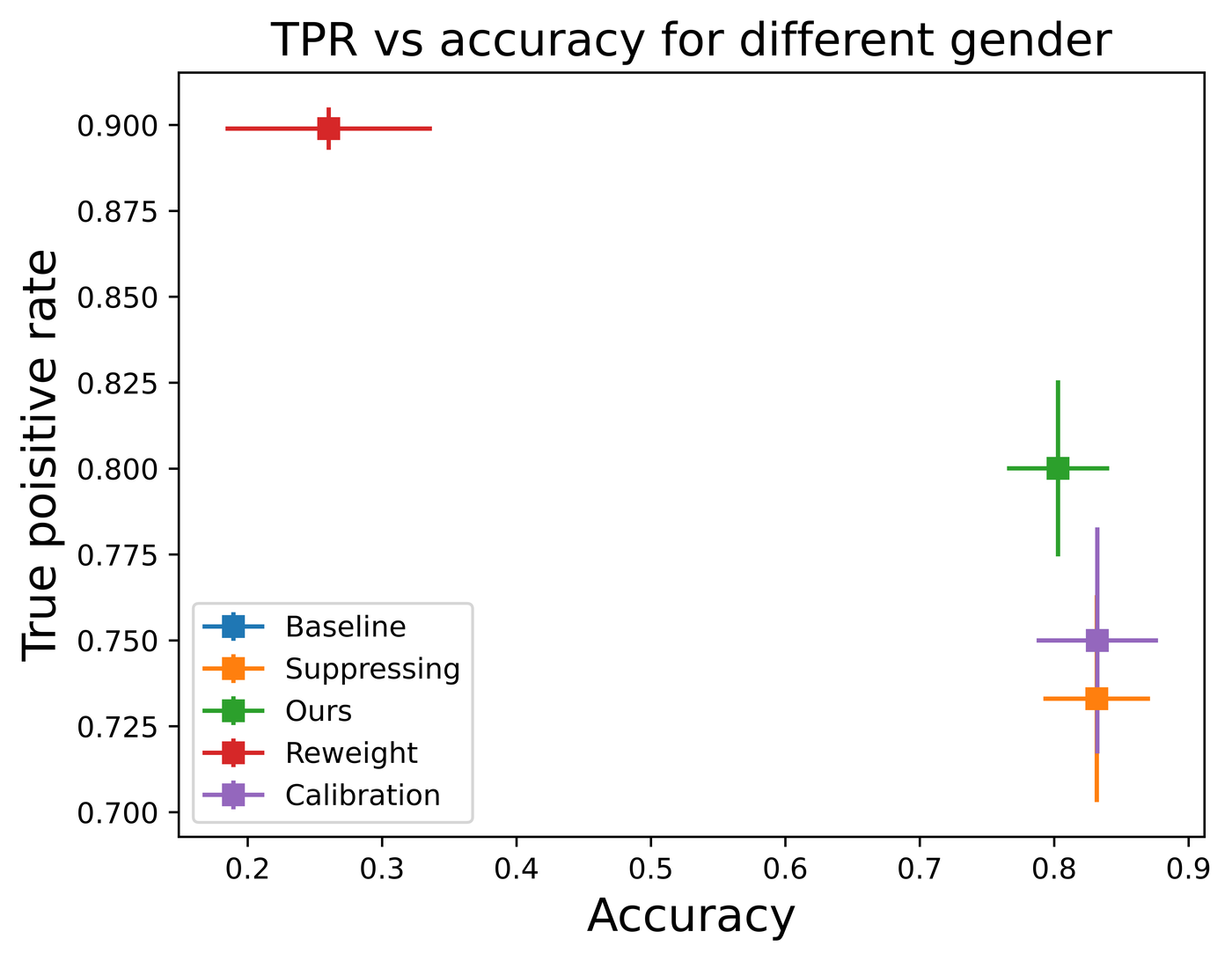}\caption{TPR vs. accuracy: gender}\end{subfigure}\hfill
\begin{subfigure}{0.32\textwidth}\includegraphics[width=\linewidth]{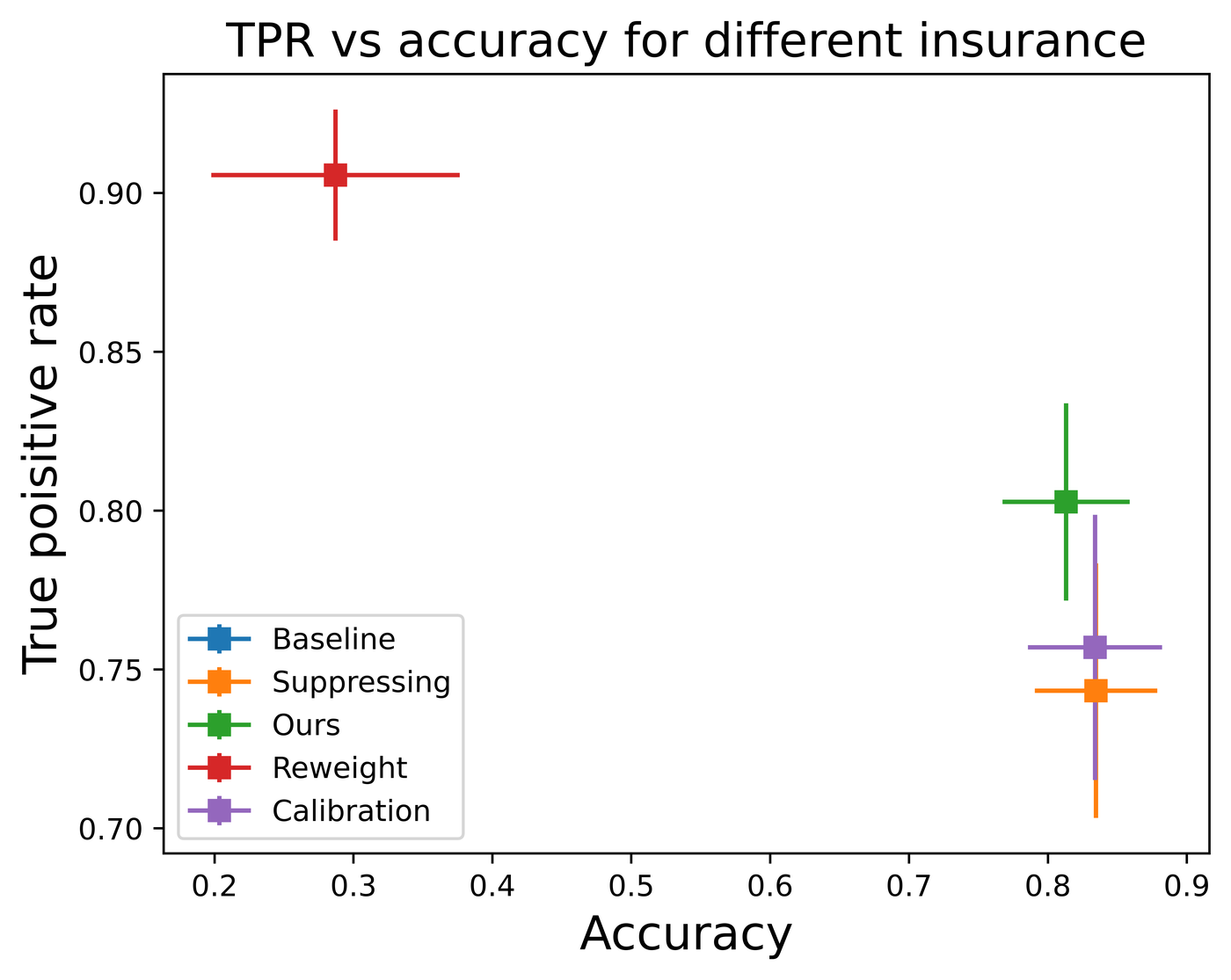}\caption{TPR vs. accuracy: insurance}\end{subfigure}
\caption{Predictive utility and sensitivity support different assessments of reweighting. High accuracy/AUROC (top) coexists with low sensitivity (bottom).}
\label{fig:tradeoff}
\end{figure}

The metric dependence remains visible when predictive utility is considered explicitly (Figure~\ref{fig:tradeoff}). Reweighting appears favorable when accuracy and AUROC are considered together, yet shifts to a high-accuracy/low-sensitivity region when TPR is plotted against accuracy. This reflects the different quantities captured by the metrics: AUROC measures discrimination across thresholds, whereas TPR and FPR characterize errors at the selected operating point. Thus, strong AUROC or accuracy does not necessarily imply clinically acceptable error behavior, while a very low FPR may still be undesirable when accompanied by a substantial loss in sensitivity. Taken together, Figures~\ref{fig:tprfpr} and~\ref{fig:tradeoff} show that the \emph{interventions are better characterized by their error profiles than by a single scalar ranking}; in mortality prediction, reporting only discrimination, accuracy, or one component of equalized odds can \textit{lead to materially different model-selection conclusions}.

\subsection{RQ2: marginal evaluation can conceal intersectional error profiles}

The fine-tuned predictions contain all 30 observed ethnicity--gender--insurance combinations. Intersection sizes range from 9 to 817 patients (median 45.5), with 1--101 mortality-positive cases per intersection (median 5). Because estimates for sparsely represented intersections can be highly uncertain, the main comparison focuses on better-supported subgroups with at least 100 patients and at least five mortality-positive cases; Appendix~\ref{app:intersectional} reports intersections with $N_{+}\geq5$.

The marginal results can obscure substantial variation within these larger subgroups. For Black patients, the marginal sensitivity is $0.805$ ($N=472$, 41 mortality-positive cases). Among the larger intersections, Black female patients with Medicare insurance have sensitivity $0.944$, FPR $0.116$, and AUROC $0.946$ ($N=113$, 18 positives), whereas Black female patients in the Other insurance category have sensitivity $0.400$, FPR $0.188$, and AUROC $0.774$ ($N=106$, 5 positives). Although the latter subgroup is not small overall, its sensitivity estimate is imprecise because it contains only five mortality-positive cases. Nevertheless, the contrast illustrates that the marginal Black sensitivity can mask substantially different error profiles across constituent intersections.

The same effect is visible in groups with stronger positive-event support. For patients in the ``Other'' ethnicity category, the marginal FPR is $0.282$ ($N=666$, 116 positives), whereas four larger three-way intersections ($N=117$--207, with 24--27 positives each) have FPRs ranging from $0.211$ to $0.371$. White patients provide a further example: the marginal group contains $2{,}997$ patients and 304 positives, with sensitivity $0.747$ and FPR $0.206$, while four Medicare/Other intersections containing 543--817 patients and 41--101 positives have sensitivities between $0.683$ and $0.782$ and FPRs between $0.153$ and $0.230$. Thus, the discrepancy between marginal and intersectional evaluation is not confined to very small demographic subgroups.

\begin{figure}[t]
\centering
\includegraphics[width=0.92\textwidth]{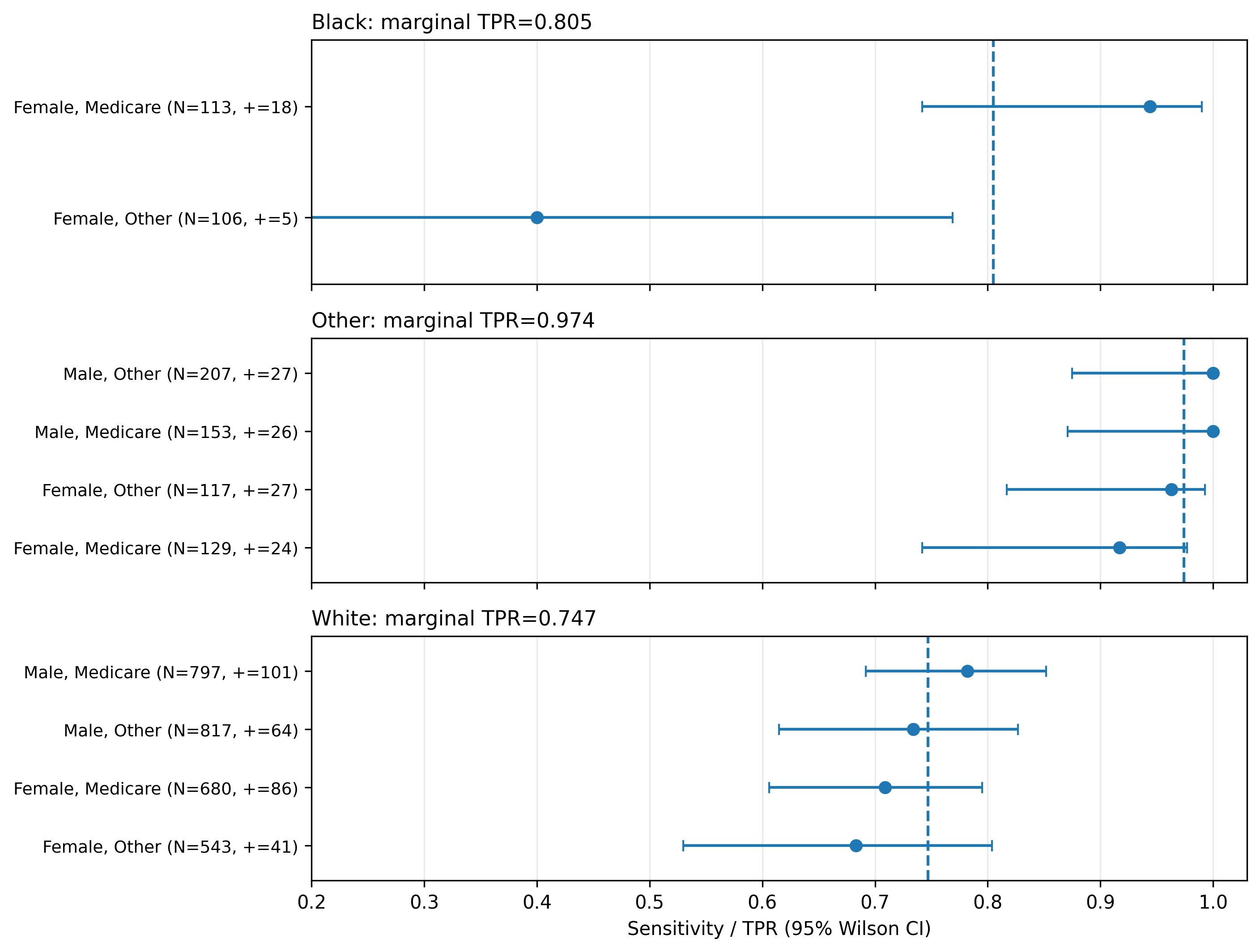}
\caption{Ethnicity-marginal sensitivity (dashed line) and sensitivity estimates for larger nested ethnicity--gender--insurance intersections. Main-text points include intersections with $N\geq100$ and at least five mortality-positive cases; horizontal bars show 95\% Wilson intervals. Dashed lines denote marginal sensitivity from the IBHFT predictions used for the intersectional analysis. Appendix~\ref{app:intersectional} reports intersections with $N_{+}\geq5$.}
\label{fig:large_intersections}
\end{figure}

Figure~\ref{fig:large_intersections} summarizes sensitivity for these larger, better-supported intersections. Across all 30 intersections, sensitivity ranges from $0.333$ to $1.000$, FPR from $0.083$ to $0.667$, and AUROC from $0.542$ to $1.000$, although the extremes partly reflect sparsely represented subgroups. Overall, the results show that demographic resolution can materially affect the error profile revealed by an evaluation: \emph{marginal summaries provide useful population-level estimates but can mask heterogeneity among their constituent intersections.}

\section{Discussion}
RQ1 shows that the evaluated interventions exhibit distinct fairness--utility profiles rather than a consistent ranking across metrics. Reweighting, for example, combines strong accuracy/AUROC and very low FPR with substantially reduced sensitivity, whereas IBHFT generally lowers FPR while retaining sensitivity closer to the baseline. Suppression and calibration produce less pronounced trade-offs. These differences are important in mortality prediction, where false-positive and false-negative errors have different implications.

IBHFT specifically targets demographic representation during adaptation without conditioning on mortality labels or directly constraining error rates. This differs from group--label balancing and outcome-conditioned augmentation \citep{mao2023lastlayer,theodorou2025fairplay}. Demographic representation, outcome prevalence, and error-rate parity are therefore distinct intervention targets, particularly when subgroup base rates differ \citep{chouldechova2017fair,pleiss2017fairness}.

Related MIMIC-IV studies reinforce the importance of both intervention target and evaluation design. \citet{lett2025intersectional} show that optimizing ethnoracial--gender intersections can yield different outcomes from marginal debiasing, while \citet{yoon2026singleaxis} report cross-axis and metric-dependent effects when interventions target race or gender separately. Our study complements these findings by jointly balancing ethnicity--gender--insurance intersections and evaluating the resulting behavior at both marginal and intersectional levels.

RQ2 shows that this evaluation resolution is itself consequential: marginal estimates can conceal substantially different intersectional error profiles. This is consistent with prior intersectional and cross-axis analyses \citep{lett2025intersectional,yoon2026singleaxis} and highlights the importance of reporting subgroup support and uncertainty alongside intersectional estimates. 

\paragraph{Key takeaways.}
\begin{itemize}[nosep, leftmargin=*]
    \item[-] \textbf{Metric choice matters.} Accuracy/AUROC, sensitivity, and FPR can support different assessments of the same fairness intervention.
    
    \item[-] \textbf{Demographic resolution matters.} Marginal summaries can conceal heterogeneous error profiles within intersectional subgroups.
    
    \item[-] \textbf{The intervention target should be explicit.} Demographic balancing, demographic--outcome balancing, and error-rate optimization should not be treated as interchangeable fairness objectives.
\end{itemize}

A deployment-oriented conclusion additionally requires a clinically meaningful decision threshold or an explicit cost structure for false-negative and false-positive errors. Without such information, the evidence supports comparison of the interventions' error profiles rather than a universal ranking of which method is preferable.

\section{Limitations}

This study uses retrospectively collected EHR data from a single health system; therefore, the generalizability of the findings to other populations, healthcare settings, and more recent clinical contexts, as well as their real-world clinical benefit, remains to be established. A useful extension would be to compare demographic-only balancing directly with demographic--outcome balancing and with variants that incorporate an explicit fairness regularizer at the intersectional subgroup level. Such experiments would help disentangle how the balancing target and regularization choice shape fairness--utility trade-offs across different demographic intersections. 
Finally, ethnicity and insurance are broad EHR categories that do not capture the full heterogeneity of the patient population; in particular, insurance type should not be interpreted as a direct proxy for socioeconomic status.

\section{Conclusion}

Using ICU mortality prediction on MIMIC-IV as a case study, we examined how evaluation choices affect the conclusions drawn about fairness interventions. The adaptation strategy balances observed ethnicity--gender--insurance representation during classifier-head fine-tuning without conditioning on mortality labels, thereby separating demographic representation balancing from outcome balancing. The resulting model comparisons are metric-dependent: high accuracy/AUROC or low FPR can coexist with poor sensitivity, so no single favorable metric is sufficient to characterize the clinical error profile. Evaluation resolution also matters. Marginal ethnicity, gender, and insurance summaries conceal materially different three-way subgroup behavior, including among better-supported intersections. These findings motivate clinical fairness evaluations that report complementary utility and error metrics, match subgroup resolution to the intervention being assessed, and make the statistical support for fine-grained subgroup estimates explicit.

\bibliographystyle{plainnat}
\bibliography{references}

@article{yoon2026singleaxis,
  title   = {Single-Axis Fairness Interventions Produce Asymmetric Cross-Axis Effects in Clinical Prediction},
  author  = {Yoon, Keun Young and Kwak, Hyunjung Gloria},
  journal = {medRxiv},
  year    = {2026},
  doi     = {10.64898/2026.06.16.26355120},
  note    = {Preprint}
}

@article{johnson2023mimiciv,
  title={MIMIC-IV, a freely accessible electronic health record dataset},
  author={Johnson, Alistair E. W. and Bulgarelli, Lucas and Shen, Lu and Gayles, Alvin and Shammout, Ayad and Horng, Steven and Pollard, Tom J. and Hao, Sicheng and Moody, Benjamin and Gow, Brian and Lehman, Li-wei H. and Celi, Leo A. and Mark, Roger G.},
  journal={Scientific Data},
  volume={10},
  pages={1},
  year={2023},
  doi={10.1038/s41597-022-01899-x}
}

@inproceedings{gupta2022pipeline,
  title={An Extensive Data Processing Pipeline for MIMIC-IV},
  author={Gupta, Mehak and Gallamoza, Brennan and Cutrona, Nicolas and Dhakal, Pranjal and Poulain, Raphael and Beheshti, Rahmatollah},
  booktitle={Machine Learning for Health},
  series={Proceedings of Machine Learning Research},
  volume={193},
  pages={311--325},
  year={2022},
  publisher={PMLR}
}

@article{meng2022mimicif,
  title={MIMIC-IF: Interpretability and Fairness Evaluation of Deep Learning Models on MIMIC-IV Dataset},
  author={Meng, Chuizheng and Trinh, Loc and Xu, Nan and Liu, Yan},
  journal={Scientific Reports},
  volume={12},
  pages={7166},
  year={2022},
  doi={10.1038/s41598-022-11012-2}
}

@inproceedings{mao2023lastlayer,
  title={Last-Layer Fairness Fine-tuning is Simple and Effective for Neural Networks},
  author={Mao, Yuzhen and Deng, Zhun and Yao, Huaxiu and Ye, Ting and Kawaguchi, Kenji and Zou, James},
  booktitle={ICML 2023 Workshop on Spurious Correlations, Invariance, and Stability},
  year={2023},
  url={https://openreview.net/forum?id=wrmynnlrTI}
}

@inproceedings{hardt2016equality,
  title={Equality of Opportunity in Supervised Learning},
  author={Hardt, Moritz and Price, Eric and Srebro, Nathan},
  booktitle={Advances in Neural Information Processing Systems},
  volume={29},
  year={2016}
}

@inproceedings{pleiss2017fairness,
  title={On Fairness and Calibration},
  author={Pleiss, Geoff and Raghavan, Manish and Wu, Felix and Kleinberg, Jon and Weinberger, Kilian Q.},
  booktitle={Advances in Neural Information Processing Systems},
  volume={30},
  year={2017}
}

@article{chouldechova2017fair,
  title={Fair prediction with disparate impact: A study of bias in recidivism prediction instruments},
  author={Chouldechova, Alexandra},
  journal={Big Data},
  volume={5},
  number={2},
  pages={153--163},
  year={2017},
  doi={10.1089/big.2016.0047}
}

@article{chen2023algorithmic,
  title={Algorithmic fairness in artificial intelligence for medicine and healthcare},
  author={Chen, Richard J. and Wang, Judy J. and Williamson, Drew F. K. and Chen, Tiffany Y. and Lipkova, Jana and Lu, Ming Y. and Sahai, Sharifa and Mahmood, Faisal},
  journal={Nature Biomedical Engineering},
  volume={7},
  pages={719--742},
  year={2023},
  doi={10.1038/s41551-023-01056-8}
}

@article{celi2022sources,
  title={Sources of bias in artificial intelligence that perpetuate healthcare disparities---A global review},
  author={Celi, Leo Anthony and Cellini, Jacqueline and Charpignon, Marie-Laure and Dee, Edward Christopher and Dernoncourt, Franck and Eber, Rene and Mitchell, William Greig and Moukheiber, Lama and Schirmer, Julian and Situ, Julia and others},
  journal={PLOS Digital Health},
  volume={1},
  number={3},
  pages={e0000022},
  year={2022},
  doi={10.1371/journal.pdig.0000022}
}

@article{liu2025scoping,
  title={A scoping review and evidence gap analysis of clinical AI fairness},
  author={Liu, Mingxuan and Ning, Yilin and Teixayavong, Salinelat and Liu, Xiaoxuan and Mertens, Mayli and Shang, Yuqing and Li, Xin and Miao, Di and Liao, Jingchi and Xu, Jie and others},
  journal={npj Digital Medicine},
  volume={8},
  pages={360},
  year={2025},
  doi={10.1038/s41746-025-01667-2}
}

@article{obermeyer2019racial,
  title={Dissecting racial bias in an algorithm used to manage the health of populations},
  author={Obermeyer, Ziad and Powers, Brian and Vogeli, Christine and Mullainathan, Sendhil},
  journal={Science},
  volume={366},
  number={6464},
  pages={447--453},
  year={2019},
  doi={10.1126/science.aax2342}
}

@article{lett2025intersectional,
  title={Intersectional and Marginal Debiasing in Prediction Models for Emergency Admissions},
  author={Lett, Elle and Shahbandegan, Shakiba and Barak-Corren, Yuval and Fine, Andrew M. and La Cava, William G.},
  journal={JAMA Network Open},
  volume={8},
  number={5},
  pages={e2512947},
  year={2025},
  doi={10.1001/jamanetworkopen.2025.12947}
}

@article{theodorou2025fairplay,
  title={Improving medical machine learning models with generative balancing for equity and excellence},
  author={Theodorou, Brandon and Danek, Benjamin and Tummala, Venkat and Kumar, Shivam Pankaj and Malin, Bradley and Sun, Jimeng},
  journal={npj Digital Medicine},
  volume={8},
  pages={100},
  year={2025},
  doi={10.1038/s41746-025-01438-z}
}

@misc{zhu2023m3fair,
  title={M$^3$Fair: Mitigating Bias in Healthcare Data through Multi-Level and Multi-Sensitive-Attribute Reweighting Method},
  author={Zhu, Yinghao and An, Jingkun and Zhou, Enshen and An, Lu and Gao, Junyi and Li, Hao and Feng, Haoran and Hou, Bo and Tang, Wen and Pan, Chengwei and Ma, Liantao},
  year={2023},
  eprint={2306.04118},
  archivePrefix={arXiv},
  primaryClass={cs.LG}
}

@inproceedings{chen2024multiple,
  title={Fairness Improvement with Multiple Protected Attributes: How Far Are We?},
  author={Chen, Zhenpeng and Zhang, Jie M. and Sarro, Federica and Harman, Mark},
  booktitle={Proceedings of the 46th IEEE/ACM International Conference on Software Engineering},
  year={2024},
  doi={10.1145/3597503.3639083}
}

@inproceedings{deng2023orthogonal,
  title={On Fairness of Medical Image Classification with Multiple Sensitive Attributes via Learning Orthogonal Representations},
  author={Deng, Wenlong and Zhong, Yuan and Dou, Qi and Li, Xiaoxiao},
  booktitle={Information Processing in Medical Imaging},
  year={2023}
}

\appendix
\section{Adaptation procedure}
\label{app:procedure}

For completeness, Algorithm~\ref{alg:ibhft} gives the adaptation procedure used in the experiments.

\begin{algorithm}[t]
\caption{Intersectionally balanced head fine-tuning}
\label{alg:ibhft}
\begin{algorithmic}[1]
\Require Training set $D_{\mathrm{tr}}$, demographic attributes $(E,G,I)$, trained base network $f_{\theta}$
\State Form observed cells $\mathcal{C}=\{(e,g,i)\}$ from $D_{\mathrm{tr}}$
\State $n_{\min}\gets \min_{c\in\mathcal{C}} |D_c|$
\For{each $c\in\mathcal{C}$}
    \State $S_c\gets$ sample $n_{\min}$ records from $D_c$ without replacement
\EndFor
\State $D_{\mathrm{bal}}\gets\bigcup_{c\in\mathcal{C}} S_c$
\State Freeze the base network through its 128-dimensional representation
\State Replace the original output layer with an MLP head $128\!\rightarrow\!512\!\rightarrow\!256\!\rightarrow\!1$
\State Optimize the head on $D_{\mathrm{bal}}$ using the mortality classification loss
\end{algorithmic}
\end{algorithm}

\section{Intersectional subgroup results}

\label{app:intersectional}

\begin{figure}[h]
\centering
\includegraphics[width=0.82\textwidth]{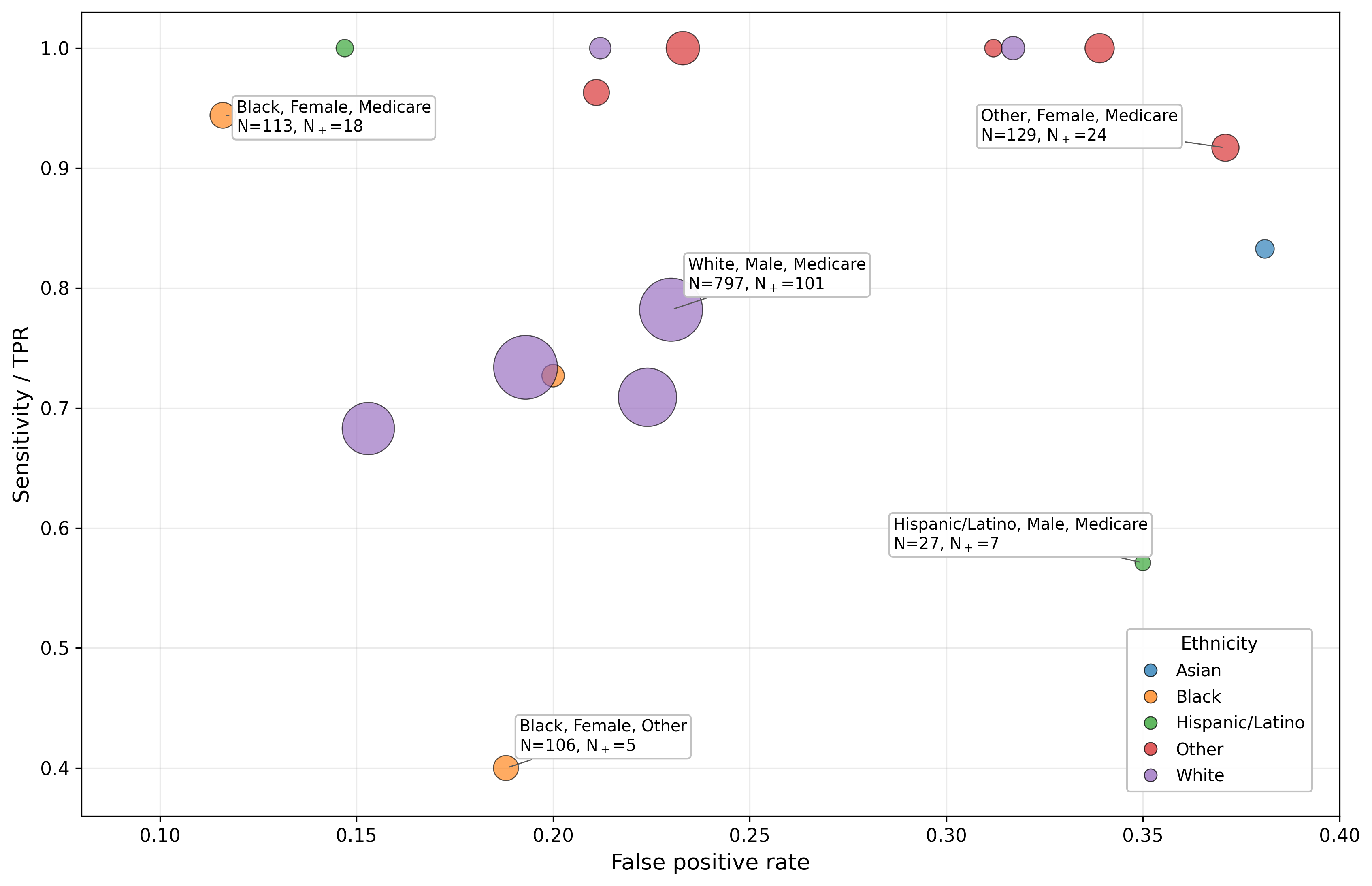}
\caption{TPR--FPR profiles for ethnicity--gender--insurance intersections with at least five mortality-positive cases ($N_+\geq5$). Marker size scales with subgroup size.}
\label{fig:appendix_tprfpr}
\end{figure}

\begin{figure}[h]
\centering
\includegraphics[width=0.90\textwidth]{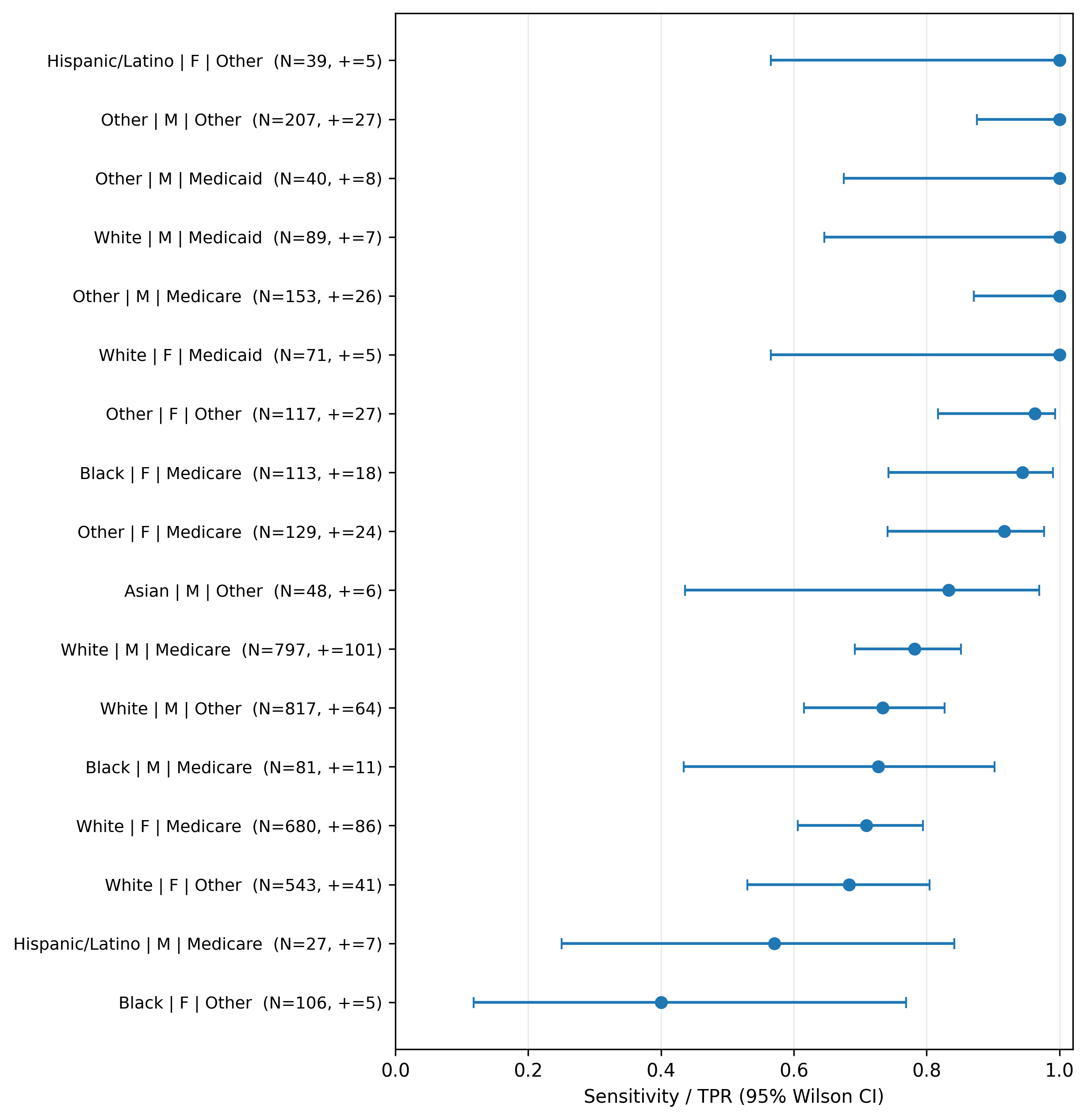}
\caption{Sensitivity estimates with 95\% Wilson intervals for ethnicity--gender--insurance intersections with at least five mortality-positive cases ($N_+\geq5$).}
\label{fig:appendix_ci}
\end{figure}

Table~\ref{tab:intersectional_full} reports ethnicity--gender--insurance intersections with at least five mortality-positive cases for the IBHFT predictions at threshold 0.5. The table summarizes subgroup support and performance, while Figure~\ref{fig:appendix_tprfpr} shows the corresponding TPR--FPR profiles and Figure~\ref{fig:appendix_ci} reports sensitivity with 95\% Wilson intervals. The main-text analysis further focuses on larger, better-supported intersections.

\begin{table}[h]
\centering
\scriptsize
\setlength{\tabcolsep}{2.8pt}
\resizebox{\textwidth}{!}{%
\begin{tabular}{lllrrrrrrrr}
\toprule
Ethnicity & Gender & Insurance & $N$ & $N_+$ & Prev. & TPR & FPR & Spec. & Acc. & AUROC \\
\midrule
Asian & M & Other & 48 & 6 & 0.125 & 0.833 & 0.381 & 0.619 & 0.646 & 0.877 \\
Black & F & Medicare & 113 & 18 & 0.159 & 0.944 & 0.116 & 0.884 & 0.894 & 0.946 \\
Black & F & Other & 106 & 5 & 0.047 & 0.400 & 0.188 & 0.812 & 0.792 & 0.774 \\
Black & M & Medicare & 81 & 11 & 0.136 & 0.727 & 0.200 & 0.800 & 0.790 & 0.800 \\
Hispanic/Latino & F & Other & 39 & 5 & 0.128 & 1.000 & 0.147 & 0.853 & 0.872 & 0.965 \\
Hispanic/Latino & M & Medicare & 27 & 7 & 0.259 & 0.571 & 0.350 & 0.650 & 0.630 & 0.750 \\
Other & F & Medicare & 129 & 24 & 0.186 & 0.917 & 0.371 & 0.629 & 0.682 & 0.829 \\
Other & F & Other & 117 & 27 & 0.231 & 0.963 & 0.211 & 0.789 & 0.829 & 0.895 \\
Other & M & Medicaid & 40 & 8 & 0.200 & 1.000 & 0.312 & 0.688 & 0.750 & 0.945 \\
Other & M & Medicare & 153 & 26 & 0.170 & 1.000 & 0.339 & 0.661 & 0.719 & 0.909 \\
Other & M & Other & 207 & 27 & 0.130 & 1.000 & 0.233 & 0.767 & 0.797 & 0.956 \\
White & F & Medicaid & 71 & 5 & 0.070 & 1.000 & 0.212 & 0.788 & 0.803 & 0.885 \\
White & F & Medicare & 680 & 86 & 0.126 & 0.709 & 0.224 & 0.776 & 0.768 & 0.839 \\
White & F & Other & 543 & 41 & 0.076 & 0.683 & 0.153 & 0.847 & 0.834 & 0.837 \\
White & M & Medicaid & 89 & 7 & 0.079 & 1.000 & 0.317 & 0.683 & 0.708 & 0.913 \\
White & M & Medicare & 797 & 101 & 0.127 & 0.782 & 0.230 & 0.770 & 0.772 & 0.862 \\
White & M & Other & 817 & 64 & 0.078 & 0.734 & 0.193 & 0.807 & 0.802 & 0.862 \\
\bottomrule
\end{tabular}%
}
\caption{Three-way intersectional audit for subgroups with at least five mortality-positive cases ($N_+\geq5$). Point estimates should be interpreted together with subgroup support; Figure~\ref{fig:appendix_ci} reports Wilson intervals for sensitivity.}
\label{tab:intersectional_full}
\end{table}

\end{document}